\documentclass{article}

\usepackage{arxiv}

\usepackage[utf8]{inputenc} 
\usepackage[T1]{fontenc}    
\usepackage{hyperref}       
\usepackage{url}            
\usepackage{booktabs}       
\usepackage{amsfonts}       
\usepackage{nicefrac}       
\usepackage{microtype}      
\usepackage{lipsum}
\usepackage{graphicx}
\usepackage{caption}
\usepackage{subcaption}
\usepackage{tikz}
\usetikzlibrary{shapes.geometric}
\graphicspath{ {./images/} }

\definecolor{fc}{HTML}{1E90FF}
\definecolor{h}{HTML}{228B22}
\definecolor{bias}{HTML}{87CEFA}
\definecolor{noise}{HTML}{8B008B}
\definecolor{convblock}{HTML}{FFA500}
\definecolor{pool}{HTML}{B22222}
\definecolor{up}{HTML}{B22222}
\definecolor{view}{HTML}{FFFFFF}
\definecolor{bn}{HTML}{FFD700}
\tikzset{fc/.style={black,draw=black,fill=fc,rectangle,minimum height=1cm}}
\tikzset{h/.style={black,draw=black,fill=h,rectangle,minimum height=1cm}}
\tikzset{bias/.style={black,draw=black,fill=bias,rectangle,minimum height=1cm}}
\tikzset{noise/.style={black,draw=black,fill=noise,rectangle,minimum height=1cm}}
\tikzset{convblock/.style={black,draw=black,fill=convblock,rectangle,minimum height=1cm}}
\tikzset{pool/.style={black,draw=black,fill=pool,rectangle,minimum height=1cm}}
\tikzset{up/.style={black,draw=black,fill=up,rectangle,minimum height=1cm}}
\tikzset{view/.style={black,draw=black,fill=view,rectangle,minimum height=1cm}}
\tikzset{data/.style={black,draw=black,fill=view,ellipse,minimum height=1cm}}
\tikzset{bn/.style={black,draw=black,fill=bn,rectangle,minimum height=1cm}}

\title{Modelling nonlinear dependencies in the latent space of inverse scattering}

\author{
 Juliusz Ziomek\\
  School of Electronics and Computer Science \\
  University of Southampton \\
  \AND
  Katayoun Farrahi \\
  School of Electronics and Computer Science \\
  University of Southampton \\
}

\begin{document}
\maketitle
\begin{abstract}
The problem of inverse scattering proposed by Angles and Mallat in 2018, concerns training a deep neural network to invert the scattering transform applied to an image. After such a network is trained, it can be used as a generative model given that we can sample from the distribution of principal components of scattering coefficients.
 For this purpose, Angles and Mallat simply use samples from independent Gaussians. However, as shown in this paper, the distribution of interest can actually be very far from normal and non-negligible dependencies might exist between different coefficients. This motivates using models for this distribution that allow for non-linear dependencies between variables. Within this paper, two such models are explored, namely a Variational AutoEncoder and a Generative Adversarial Network. We demonstrate the results obtained can be extremely realistic on some datasets and look better than those produced by Angles and Mallat. The conducted meta-analysis also shows a clear practical advantage of such constructed generative models in terms of the efficiency of their training process compared to existing generative models for images.
\end{abstract}

\section{Introduction}
Scattering networks have been proposed by Bruna and Mallat \cite{MallatScattering}. They are a hard-coded transformation designed to be stable with respect to deformations. This implies that a small deformation (which can be non-linear) changing the input should produce a relatively small change in the feature space of its representation. In the paper where scattering networks for image data are proposed, Bruna and Mallat show that a classifier trained on the representation produced by scattering networks on MNIST dataset, outperforms convolutional neural networks in case when limited training data is available. Later, in 2018, Angles and Mallat \cite{MallatInverse} propose a generative model using the representation produced by scattering networks. In their model, this representation is first obtained for the entire training set, after which Principal Component Analysis is applied to reduce the dimensionality of this representation to 512 features. This representation is then whitened to remove all linear dependencies between features. A deep neural network is subsequently trained to invert this operation and reconstruct an image from the principal components of the scattering network's representation. Such a trained neural network can then be used as a generative model and produce artificial images given artificial principal coefficients. To generate artificial principal coefficients, Angles and Mallat use a distribution of independent Gaussians. They argue that as the scale of the scattering network grows, the distribution of the principal coefficients should approach a Gaussian. This follows directly from the central limit theorem under the assumption that with enough distance pixels will become independent of each other. However, in scattering networks we usually use relatively small scale and in the case of experiments of Angles and Mallat, the averaging occurred over windows of size 16 by 16 pixels while working with images of size 128 by 128. In natural images of such size, locations separated by less than 16 pixels might be highly dependent, meaning that the distribution of interest might be relatively far from Gaussian. The empirical evidence agrees with this conclusion as the results shown by Angles and Mallat do not closely resemble the training distribution. The main aim of this paper is to investigate whether using artificial principal coefficients sampled from models capable of including nonlinear dependencies between features can improve those results.
\section{Related Work}
\subsection{Variational Autoencoders}
Variational autoencoders (VAE) are a class of autoencoders based on the framework developed by Kingma and Welling \cite{VAE}. They force the model to learn a latent representation with a distribution close to independent Gaussians. The cost function used in these models is a sum of the reconstruction loss which measures how well the image is reconstructed from its representation and a KL-divergence between the distribution of latent space and independent Gaussians. Later, disentangling VAEs \cite{betaVAE} were proposed, which additionally weigh the KL-divergence term by a $\beta$ constant.
\subsection{Generative Adversarial Networks}
Goodfellow et al \cite{goodfellow2014GAN} propose an adversarial process in which two models are trained simultaneously, the generator $G$ and discriminator $D$. The task of the generator is to convert $H$-dimensional, random noise vector $\mathbf{z} \sim p(\mathbf{z})$ to a datapoint resembling those coming from given data distribution $p(\mathbf{x})$. The task of the discriminator is to differentiate between the true datapoints $\mathbf{x}$ and the fake datapoints $\hat{\mathbf{x}}$ generated by $G$. Since their invention, GANs have dominated the landscape of deep generative models and established the state of art on many tasks.
\subsection{Other generative models utilising scattering networks}
Oyallon et al \cite{eduardohybrid} propose a different approach to use scattering networks in generative models than the one used by Angles and Mallat \cite{MallatInverse}. They use a GAN in the coefficient space to generate artificial scattering coefficients (generator produces artificial scattering coefficients and discriminator tries to distinguish them from real scattering coefficients). However, they work directly with high-dimensional scattering coefficients and not the low-dimensional principal coefficients. They also map the coefficients to the image domain via expensive numerical reconstructions rather than via a neural network. 
\section{Experiment Setup}
The experiments in the following sections were conducted on two datasets:  MNIST hand-written digits and CelebA \cite{liu2015CelebA} containing photos of faces of celebrities.
For the MNIST dataset, whole images of shape ($28$,$28$) are used. For CelebA dataset, centre-crops of shape (128,128) are used. Before any further work could be done, the representations of images in form of whitened scattering coefficients had to be obtained. The scattering network scale of $J=4$ was used for CelebA. The scale of $J=2$ was used for MNIST, as this dataset has much smaller images. The experimental setup for CelebA was chosen to faithfully reproduce the work of  \cite{MallatInverse}, as they use the same image shapes, same scale of scattering networks and same size of train sets (65,536). \\ \\
When it comes to training the networks mapping representations to the image domain, we use the same architecture as  \cite{MallatInverse} for CelebA and an architecture with reduced capacity for MNIST, where we set the filter size for all convolutional layers to 3 and the size of the last hidden layer to 32, where previous layers have sizes following geometric progression with a ratio of $1/2$.
\section{Investigating nonlinear dependencies in the latent space}
\subsection{Visualisation of principal components}
\begin{figure}[h]
    \centering
    \begin{subfigure}[b]{0.3\textwidth}
        \includegraphics[width=\textwidth]{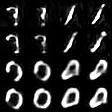}
        \caption{MNIST}
    \end{subfigure}
    \begin{subfigure}[b]{0.3\textwidth}
        \includegraphics[width=\textwidth]{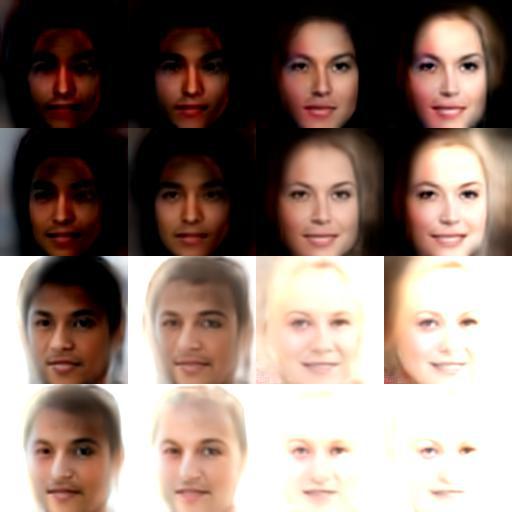}
        \caption{CelebA}
    \end{subfigure}
    \caption{Visualisation matrix of components mapped to the image domain. Each column top to bottom corresponds to increasing the value of the first coefficient. Each row from left to right corresponds to increasing the value of the second coefficient.}
    \label{fig:visual_matrix}
\end{figure}

To better understand what the principal components represent, we visualise the effect of changing the ones with the greatest variance before whitening. For each dataset considered, a "visualisation matrix" was created for the two most significant components. Such a matrix consists of 16 images that are created by mapping different vectors to the image domain. Each column corresponds to increasing the value of the first coefficient and each row to increasing the value of the second coefficient. Values of those components were chosen from a set $\{-10,-5,5,10\}$. The values of the remaining 510 components were set to $0$. The result of this process is shown in Figure \ref{fig:visual_matrix}. The visualisation matrices graphically illustrate the dependencies between the first two principal components. One can see that a given value of the first coefficient produces a realistic image for some specific values of the second coefficient, but not for others. This would suggest that those components must be "jointly realistic" to produce a realistic image, meaning that they are most likely not independent. This conclusion aligns with what is indicated by statistical tests in the next paragraph.
\subsection{Statistical tests for normality}
Direct testing for the independence of multi-dimensional continuous data is difficult and computationally infeasible. Instead, we measure how much individual distributions of principal components differ from a Gaussian. If those distributions are perfect Gaussians, then they should be completely independent, because they were whitened. If they differ significantly from a Gaussian, whitening does not guarantee independence and it is reasonable to assume there might be (nonlinear) dependencies between them. To test how likely it is that those components come from a Gaussian distribution, we conduct two statistical tests for normality: D’Agostino’s K-squared (K2) and Jarque–Bera (J-B) for each of the component for MNIST and CelebA datasets. The results are shown in Table \ref{tab:normality_tests}. At the significance level of $\alpha = 0.05$ depending on dataset and test 35-51 components (7\% - 10\%) have normality hypothesis rejected and on level of $\alpha = 0.01$ this number is equal to 12 - 24 (2\% - 4\%). This is enough to conclude that at least some components have distributions significantly differing from a Gaussian. The main question that remains is what kind of information those components will usually encode and how setting them to unlikely values will affect the reconstructed image. Within the next section we show that using more complex models to sample the principal components can have an overwhelmingly positive effect on the quality of samples generated.
\begin{table}[h]
    \centering
    \begin{tabular}{c|c|cc}
    Dataset & Test type & \multicolumn{2}{|c}{Number of comp. rejected at}\\
    & & \textbf{0.05-level} & \textbf{0.01-level}  \\
    \midrule
    MNIST & K2 & \textbf{41} & \textbf{12}\\
    & JB & \textbf{35} & \textbf{16}\\ \midrule
    CelebA & K2 & \textbf{46} & \textbf{19}\\
    & JB & \textbf{51} & \textbf{24}\\
    \end{tabular}
    
    \caption{Number of principal components for which the normality hypothesis can be rejected at a given $\alpha$-level (ie. their p-value is lower than $\alpha$). Two types of tests: D'Agostino's K-squared test (K2) and Jarque–Bera (JB) were conducted.}
    \label{tab:normality_tests}
\end{table}
\section{Introducing nonlinear models}
\subsection{Models}
The first model we use for generating artificial principal coefficients was the variational autoencoder(VAE). Since the principal components are of low-dimensionality, convolutional layers were not needed and the model was constructed using only fully connected layers. We use $H=64$ as the size of the latent space. The architecture of VAEs used for experiments is shown in Appendix \ref{VAE_arch}.
The second model we use was a GAN, and as in the case of a VAE, we use only fully connected layers. The dimensionality of the input noise vector $H$ was set to $H=64$. The exact architecture of the GAN used for experiments is shown in Appendix \ref{GAN_arch}.
\subsection{Results}
\begin{figure}[h]
    \centering
    \begin{subfigure}[b]{0.25\textwidth}
        \includegraphics[width=\textwidth]{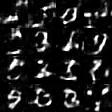}
        \caption{Gaussian Noise}
    \end{subfigure}
    \begin{subfigure}[b]{0.25\textwidth}
        \includegraphics[width=\textwidth]{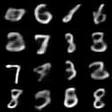}
        \caption{VAE}
    \end{subfigure}
    \begin{subfigure}[b]{0.25\textwidth}
        \includegraphics[width=\textwidth]{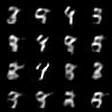}
        \caption{GAN}
    \end{subfigure}
    \caption{Results of mapping artificial principal components generated by different methods to image domain on the MNIST dataset.}
    \label{fig:MNIST}
\end{figure}
\begin{figure}[h]
    \centering
    \begin{subfigure}[b]{0.25\textwidth}
        \includegraphics[width=\textwidth]{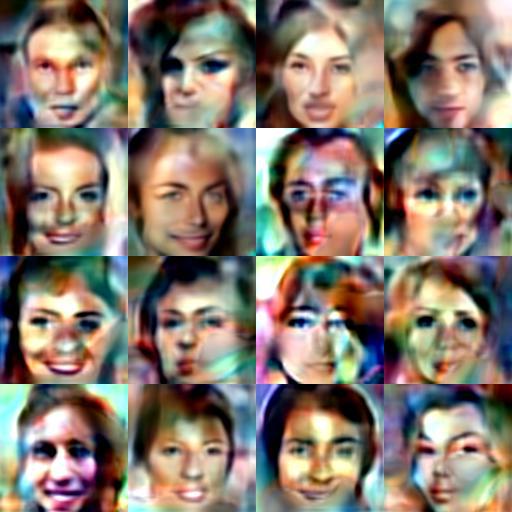}
        \caption{Gaussian Noise}
    \end{subfigure}
    \begin{subfigure}[b]{0.25\textwidth}
        \includegraphics[width=\textwidth]{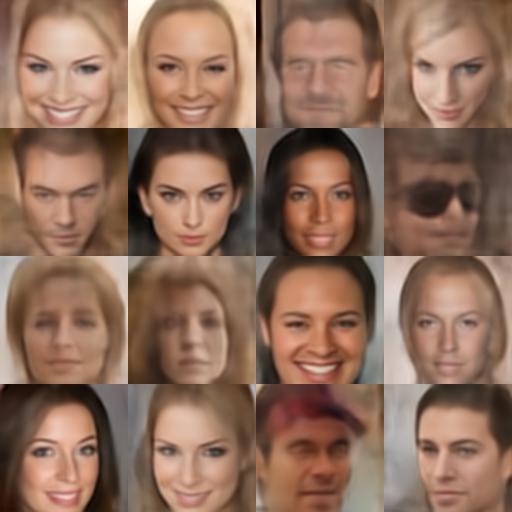}
        \caption{VAE}
    \end{subfigure}
    \begin{subfigure}[b]{0.25\textwidth}
        \includegraphics[width=\textwidth]{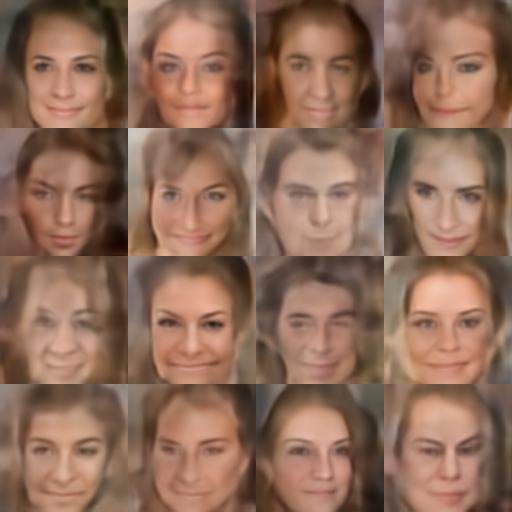}
        \caption{GAN}
    \end{subfigure}
    \caption{Results of mapping artificial principal componenets generated by different methods to image domain on the CelebA dataset.}
    \label{fig:CelebA}
\end{figure}
\begin{table}[!h]
  \centering
  \begin{tabular}{c|ccc}
  \toprule
  \textbf{Dataset} & \textbf{Used $\beta$}  & \textbf{Epochs} & \textbf{Time Elapsed}  \\
  \midrule
  MNIST & $0.001$ & $31$ & 9 mins\\
  CelebA & $0.0025$ & $96$ & 17 mins\\
  \bottomrule
  \end{tabular}
  \caption{Summary of the VAE training process for different datasets.}
  \label{Table:VAE_results}
  
\end{table}
\begin{table}[!h]
  \centering
  \begin{tabular}{c|cc}
  \toprule
  \textbf{Dataset} & \textbf{Epochs} & \textbf{Time Elapsed}  \\
  \midrule
  MNIST & $30$ & 5 mins\\
  CelebA & $38$ & 10 mins\\
  \bottomrule
  \end{tabular}
  \caption{Summary of the GAN training process for different datasets.}
  \label{Table:GAN_results}
  
\end{table}
The results of mapping artifical principal components generated by different methods to the image domain is shown in Figure \ref{fig:MNIST} for the MNIST dataset and in Figure \ref{fig:CelebA} for the CelebA dataset. Additionally, we report training times and $\beta$ values used for the VAE models in Table \ref{Table:VAE_results} and training times of GAN models in Table \ref{Table:GAN_results}.
\section{Discussion}
Both VAEs and GANs are able to outperform Gaussian Noise on MNIST dataset (compare Figure \ref{fig:MNIST} a) with b) and c)).
VAEs are also able to obtain excellent results on the CelebA dataset, where some samples are arguably indistinguishable from real people. Together with the results of statistical tests this is enough to claim that principal components containing information about important image features can have nonlinear dependencies between each other. GANs also show an improvement over generation from Gaussians, producing more consistent images, however, their improvement is not as significant as that of VAEs. GANs also generate very similar samples, which is due to a phenomenon called the "mode-collapse". However, as argued by Ian Goodfellow \cite{goodfellow2017nips}, this is a general tendency of GAN models and while it can be reduced, it cannot be completely avoided. \\\\
Possibly the most important advantage of the generative models constructed within the scope of this work is not their performance but their efficiency. Although the performance of some methods is excellent on particular datasets, it is likely that a generative model trained end-to-end will produce samples of better quality. Such an end-to-end model, however, will have one significant issue - training time. Training a deep, convolutional network often takes hours or even days. Moreover, training a generative model is a very unstable process, which is highly sensitive to hyperparameters. Therefore, such expensive training has to be repeated multiple time, before a sensible model is obtained. A tremendous advantage of the presented approach is the separation of expensive deep, convolutional network used to map to image domain from cheap and shallow networks used to generate principal components. The mapping network is trained to map a given set of inputs to corresponding outputs, and this training has no generative aspect. Such a process is straightforward to perform and multiple efficient methods have been developed to highly optimise it. Such non-generative training is also stable and not that sensitive to hyperparameters, meaning that in general, it does not have to be repeated. All of the instability lies in training a completely independent generating network, which can be a shallow, fully connected network as the principal coefficients have low dimensionality. As mentioned before, this training usually has to be repeated multiple times, however, since the network is lightweight, such repetitions are not computationally expensive. A meta-analysis was conducted to estimate typical times of generative training for different datasets. The results are shown together with methods proposed within this work in Table \ref{tab:meta}. It already shows that the total generative training time of a generative scattering model is at least an order of magnitude lower than that of a typical VAE and two orders of magnitude lower than that of a typical GAN. In a case where many different hyperparameters of the generative model have to be tested, the total time of all experiments would be significantly shorter for a generative scattering model. 
\begin{table}[h]
    \centering
    \begin{tabular}{c|c|c|c|c}
         Source & Dataset & Method & Generative  & Non-generative   \\
         & (dimensionality)& & training & training \\
         \midrule
         This project & CelebA & VAE & \textbf{17 mins} & \textbf{27 h 45 mins} \\
         & (128x128x3) & in scatter. space   &  & \\ 
         \midrule
          \cite{VAE_4x80x96x64_12h} & MR Brain & VAE & \textbf{10 h} & \textbf{-}\\
         &  (4x80x96x64) & & & \\
         \midrule
          \cite{vae_400x400_6h} & CMB & VAE & \textbf{6 h} & \textbf{-}\\
         &  (400x400) & & & \\
         \midrule
          \cite{vae_180_240_3_4_h} & Fashion img. & VAE & \textbf{4 h} & \textbf{-}\\
         &  (180x240x3) & & & \\
         \midrule
         This project & CelebA & GAN & \textbf{10 mins} & \textbf{27 h 45 mins} \\
         & (128x128x3) & in scatter. space  &  & \\ 
         \midrule
          \cite{CelebA_64x64_10h} & CelebA & GAN & \textbf{10 h} & \textbf{-}\\
         &  (64x64x3) & & & \\
         \midrule
          \cite{rivers256x256_73h} & Landscape img.& GAN & \textbf{73 h} & \textbf{-}\\
         &  (256x256x3) & & & \\
         
    \end{tabular}
    \caption{Results of the meta-analysis regarding training time of generative models.}
    \label{tab:meta}
\end{table}
\section{Conclusions}
In this paper we have provided strong arguments for the thesis that principal components of outputs of scattering networks for natural images, might have nonlinear dependencies and be not modelled accurately by independent Gaussians. We also show that using a model capable of capturing those nonlinear dependencies greatly improves the quality of images obtaining after mapping the components to image domain. We also compare our proposed method to exisiting ones and identify an improvement in terms of efficiency, as in our method the unstable generative training only concerns a very small and shallow neural network.

\bibliographystyle{unsrt}  

\bibliography{references}
\clearpage
\appendix
\section{VAE architecture} \label{VAE_arch}
\begin{figure}[h]
  \begin{subfigure}[b]{0.5\textwidth}
  \begin{tikzpicture}
    \node[data,rotate=0,minimum width=2cm] (z) at (0,-1.25) {$\mathbf{x}$};
    \node[fc,rotate=0,minimum width=6cm] (fc1) at (0,0) {FC Layer ($N$, $4H$) + ReLU };
    \node[fc,rotate=0,minimum width=6cm] (fc2) at (0,1.25) {FC Layer ($4H$, $4H$) + ReLU};
    \node[fc,rotate=0,minimum width=6cm] (fc3) at (0,2.5) {FC Layer ($4H$, $2H$) + ReLU };
    \node[fc,rotate=0,minimum width=2cm] (mean_lay) at (-2,3.75) {FC Layer ($2H$,$H$)};
    \node[fc,rotate=0,minimum width=2cm] (var_lay) at (2,3.75) {FC Layer ($2H$,$H$)};
    
    \node[data,rotate=0,minimum width=2cm] (mean) at (-2,5) {$\mathbf{\mu}$};
    \node[data,rotate=0,minimum width=2cm] (var) at (2,5) {$\log \mathbf{\sigma}^{2}$};
    
    \draw[->] (z) -- (fc1);
    \draw[->] (fc1) -- (fc2);
    \draw[->] (fc2) -- (fc3);
    \draw[->] (fc3) -- (mean_lay);
    \draw[->] (fc3) -- (var_lay);
    \draw[->] (mean_lay) -- (mean);
    \draw[->] (var_lay) -- (var);
 
  \end{tikzpicture}
  \caption{Encoder}
  \end{subfigure}
\begin{subfigure}[b]{0.5\textwidth}
  \begin{tikzpicture}
 
   \node[data,rotate=0,minimum width=2cm] (e) at (-2.5,1.25) {$\mathbf{\epsilon} \sim \mathcal{N}(0,1) $};
 
   \node[bn,rotate=0,minimum width=1cm] (mul) at (0,1.25) {$*$};
   \node[data,rotate=0,minimum width=1cm] (std) at (0,0) {$\mathbf{\sigma}$};
   \node[bn,rotate=0,minimum width=1cm] (add) at (1.25,1.25) {$+$};
   \node[fc,rotate=0,minimum width=1cm] (fc1handler) at (1.25,2.5) {};
   \node[data,rotate=0,minimum width=1cm] (mean) at (1.25,0) {$\mathbf{\mu}$};
   \node[fc,rotate=0,minimum width=6cm] (fc1) at (0,2.5) {FC Layer ($H$, $4H$) + ReLU };
   \node[fc,rotate=0,minimum width=6cm] (fc2) at (0,3.75) {FC Layer ($4H$, $4H$) + ReLU};
   \node[fc,rotate=0,minimum width=6cm] (fc3) at (0,5) {FC Layer ($4H$, $N$)};
   \node[data,rotate=0,minimum width=2cm] (z) at (0,6.25) {$\mathbf{\hat{x}}$};

    \draw[->] (e) -- (mul);
    \draw[->] (std) -- (mul);
    \draw[->] (mul) -- (add);
    \draw[->] (mean) -- (add);
    \draw[->] (add) -- (fc1handler);
    \draw[->] (fc1) -- (fc2);
    \draw[->] (fc2) -- (fc3);
    \draw[->] (fc3) -- (z);

  \end{tikzpicture}
  \caption{Decoder}
\end{subfigure}
  
  \caption{\label{Figure:VAE_architecture} Schematic of the used VAE architecture. $N$ represents the number of input features and $H$ is the size of the latent representation. $\mathbf{x}$ denotes the input.}
 \end{figure}
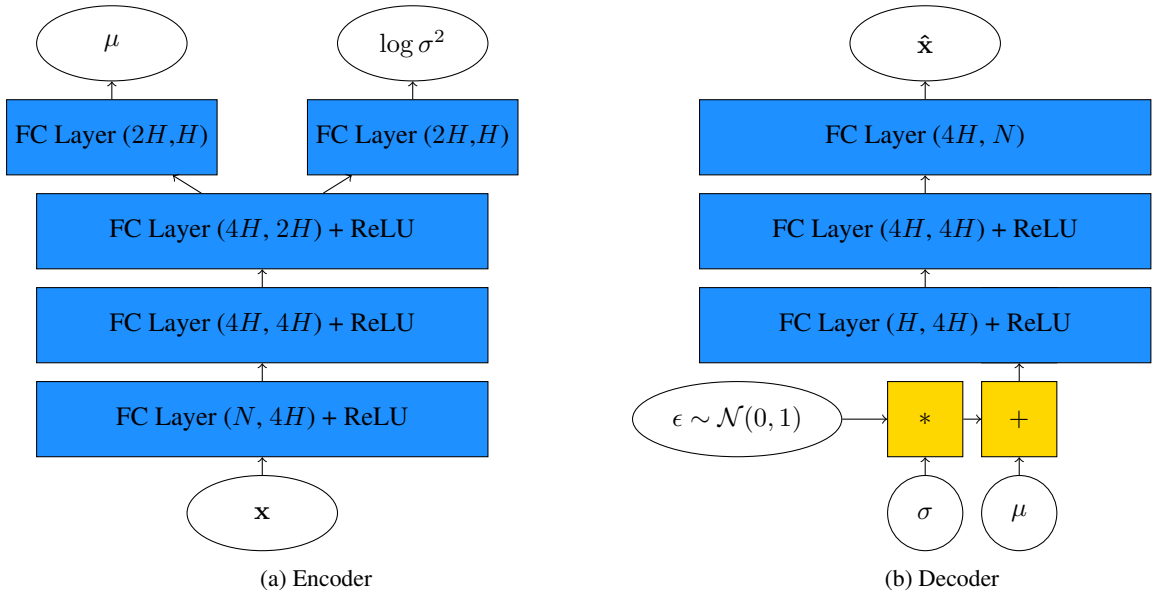
 
\clearpage

\section{GAN architecture} \label{GAN_arch}
 \begin{figure}[h]
  \begin{subfigure}[b]{0.5\textwidth}
  \begin{tikzpicture}
    \node[data,rotate=0,minimum width=2cm] (z) at (0,-1.25) {$\mathbf{z}$};
    \node[fc,rotate=0,minimum width=6.5cm] (fc1) at (0,0) {FC Layer ($H$, $2H$)};
    \node[bn,rotate=0,minimum width=6.5cm] (bn1) at (0,1.25) {Batch Norm. + ReLU};
    \node[fc,rotate=0,minimum width=6.5cm] (fc2) at (0,2.5) {FC Layer ($2H$, $4H$)};
    \node[bn,rotate=0,minimum width=6.5cm] (bn2) at (0,3.75) {Batch Norm. + ReLU};
    \node[fc,rotate=0,minimum width=6.5cm] (fc3) at (0,5) {FC Layer ($4H$, $N$) + ReLU};
    \node[data,rotate=0,minimum width=2cm] (x) at (0,6.25) {$\mathbf{x}_{fake}$};
    
    \draw[->] (z) -- (fc1);
    \draw[->] (fc1) -- (bn1);
    \draw[->] (bn1) -- (fc2);
    \draw[->] (fc2) -- (bn2);
    \draw[->] (bn2) -- (fc3);
    \draw[->] (fc3) -- (x);
 
  \end{tikzpicture}
  \caption{Generator}
  \end{subfigure}
\begin{subfigure}[b]{0.5\textwidth}
  \begin{tikzpicture}
 
    \node[data,rotate=0,minimum width=2cm] (x) at (0,-1.25) {$\mathbf{x}$};
    \node[fc,rotate=0,minimum width=6.5cm] (fc1) at (0,0) {FC Layer ($N$, $N/2$)};
    \node[bn,rotate=0,minimum width=6.5cm] (bn1) at (0,1.25) {LeakyReLU(0.2)};
    \node[fc,rotate=0,minimum width=6.5cm] (fc2) at (0,2.5) {FC Layer ($N/2$, $N/4$)};
    \node[bn,rotate=0,minimum width=6.5cm] (bn2) at (0,3.75) {Batch Norm. + LeakyReLU(0.2)};
    \node[fc,rotate=0,minimum width=6.5cm] (fc3) at (0,5) {FC Layer ($N/4$, $1$)};
    \node[bn,rotate=0,minimum width=6.5cm] (out) at (0,6.25) {Sigmoid};
    \node[data,rotate=0,minimum width=2cm] (y) at (0,7.5) {$y$};
    
    \draw[->] (x) -- (fc1);
    \draw[->] (fc1) -- (bn1);
    \draw[->] (bn1) -- (fc2);
    \draw[->] (fc2) -- (bn2);
    \draw[->] (bn2) -- (fc3);
    \draw[->] (fc3) -- (out);
    \draw[->] (out) -- (y);
 
  \end{tikzpicture}
  \caption{Discriminator}
  \end{subfigure}
  
  \caption{\label{Figure:GAN_architecture} Schematic of the used GAN architecture. $N$ represents the input size and $H$ is the dimensionality of the input, random noise vector.}
 \end{figure}
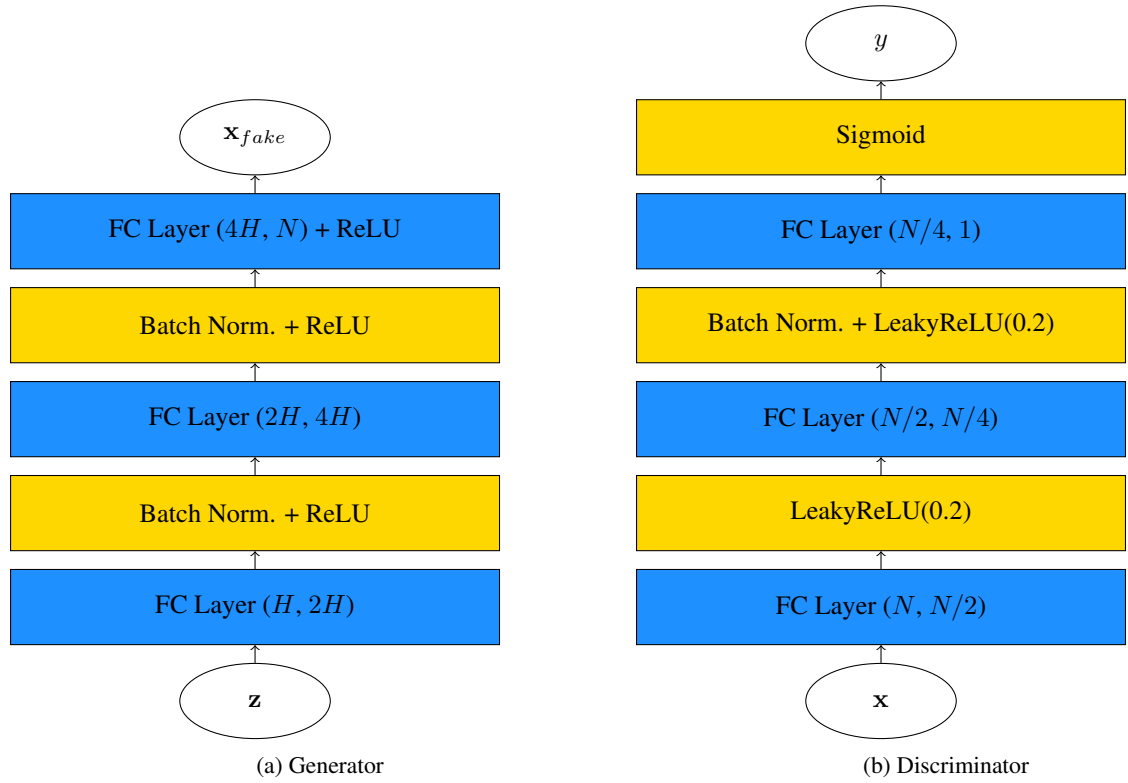
\end{document}